# Deeply-Supervised Recurrent Convolutional Neural Network for Saliency Detection

Youbao Tang, Xiangqian Wu
School of Computer Science and Technology
Harbin Institute of Technology
Harbin, 150001 China
{tangyoubao, xqwu}@hit.edu.cn

Wei Bu
Department of New Media Technologies and Arts
Harbin Institute of Technology
Harbin, 150001 China
buwei@hit.edu.cn

## ABSTRACT
This paper proposes a novel saliency detection method by developing a deeply-supervised recurrent convolutional neural network (DSRCNN), which performs a full image-to-image saliency prediction. For saliency detection, the local, global, and contextual information of salient objects is important to obtain a high quality salient map. To achieve this goal, the DSRCNN is designed based on VGGNet-16. Firstly, the recurrent connections are incorporated into each convolutional layer, which can make the model more powerful for learning the contextual information. Secondly, side-output layers are added to conduct the deeply-supervised operation, which can make the model learn more discriminative and robust features by effecting the intermediate layers. Finally, all of the side-outputs are fused to integrate the local and global information to get the final saliency detection results. Therefore, the DSRCNN combines the advantages of recurrent convolutional neural networks and deeply-supervised nets. The DSRCNN model is tested on five benchmark datasets, and experimental results demonstrate that the proposed method significantly outperforms the state-of-the-art saliency detection approaches on all test datasets.

## Keywords
Saliency detection; deeply-supervised recurrent convolutional neural network; deeply-supervised learning; recurrent connection.

## 1. INTRODUCTION
Visual saliency detection, which aims to highlight the most important object regions in an image, is an important and challenging task in computer vision and has received extensive attentions in recent years. Numerous image processing applications incorporate the visual saliency to improve their performance, such as image segmentation [1] and cropping [2], object detection [3], and image retrieval [4], etc. A large number of visual saliency detection approaches [5-37] have been proposed by exploiting different salient cues recently. They can be roughly categorized as hand-designed features based approaches and CNN based approaches.

For the hand-designed features based approaches, the local and global features are extracted from pixels [5-7, 30] and regions [8-11, 18, 19, 23, 25, 27, 32-35, 37] for saliency detection. Generally, the approaches extracted features from pixels highlight high-contrast edges instead of the salient objects, or get low contrast salient maps. That is because the extracted features are unable to reflect the complex relationship between pixels. The approaches extracted features from regions are much more effective than the ones from pixels to detect the saliency, since more sophisticated and discriminative features can be extracted from regions. Even so, the hand-designed features cannot capture the high-level and multi-scale information of salient objects.

Convolutional neural network (CNN) is powerful for high-level and multi-scale feature learning, which has been successfully used in many applications of computer vision, such as semantic segmentation [38, 39] and edge detection [40, 41]. In last year, several researchers [21, 26, 28] also propose CNN based approaches for saliency detection and get the state-of-the-art performance. That is because CNNs are able to extract more robust and discriminative features with considering the global contextual information of regions. All of these approaches [21, 26, 28] first segment images into a number of regions (such as sliding windows [28], multi-level decomposition regions [21], and superpixels [26]) and then use CNN to extract features from these regions, based on which the saliencies are estimated. Therefore, these approaches are complex and cannot generate the precise pixel-wise saliency prediction.

To overcome these problems, this work proposes a full image-to-image saliency prediction by developing a deeply-supervised recurrent convolutional neural network (DSRCNN), which is based on VGGNet-16 [42]. The DSRCNN combines the advantages of recurrent convolutional neural networks [43] and deeply-supervised nets [44] by incorporating the recurrent connections into each convolutional layer and adding side-output layers to supervise the feature learning of the intermediate layers. Therefore, an input image is fed into the DSRCNN model and a high quality salient map of the same size as the input image is directly produced after the feed-forward process. Figure 1 shows some saliency detection results of the proposed method, which are very close to the ground truths.

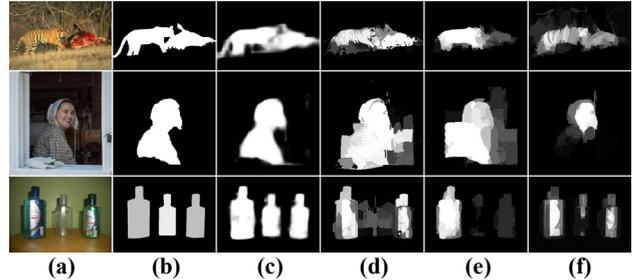

(a)   (b)   (c)   (d)   (e)   (f)

Figure 1. Three examples of saliency detection results. (a) The input images. (b) The ground truths. (c) Our detection results. (d)-(f) The salient maps detected by the state-of-the-art CNN based approaches, i.e. MC [26], LEGS [28], and MDF [21], respectively.

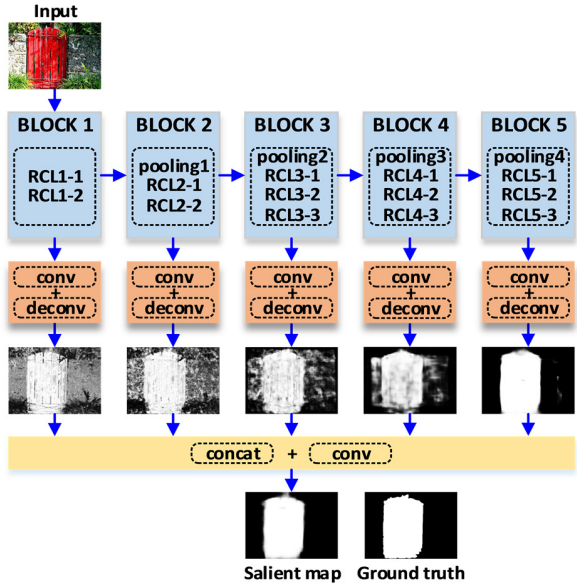

Figure 2. The architecture of the proposed deeply-supervised recurrent convolutional neural network for saliency detection.

## 2. DSRCNN FOR SALIENCY DETECTION

Recently, recurrent convolutional neural network (RCNN) has achieved state-of-the-art results of object recognition [43], and deeply-supervised nets (DSN) [44] have been successfully used in image classification [44] and edge detection [40]. RCNN has the ability to learn the contextual features, and DSN is able to learn more discriminative and robust features from local and global view. The contextual, local, and global features are important to extract a high quality salient map. Therefore, a deeply-supervised recurrent convolutional neural network (DSRCNN) is designed for saliency detection in this paper. In this section, we describe the DSRCNN in detail.

To obtain a precise saliency prediction, the CNN architecture should be deep and have multi-scale stages with different strides, so that we can learn discriminative and multi-scale features for pixels. Training such a deep network from scratch is difficult when there are few training samples. So this work chooses the VGGNet-16 [42] trained on the large-scale dataset as the pre-trained model for fine-tuning as done by [40]. The VGGNet-16 has six blocks. The first five blocks contain convolutional layers and pooling layers, and the last block contains a pooling layer and fully connected layers. The last block is removed in this work, since the pooling operation in this block makes the feature maps become too small (about $1/32$ of the input image size) to obtain fine full-size prediction and the fully connected layers are time and memory consuming. Based on the first five blocks of VGGNet-16, the deeply-supervised recurrent convolutional neural network (DSRCNN) is constructed as shown in Figure 2.

To make the model learn more contextual information, we use the recurrent convolutional layers (RCL) (as shown in Figure 3(b)) to replace the convolutional layers (as shown in Figure 3(a)) in the five blocks by incorporating the recurrent connections into each convolutional layer. Unfolding the recurrent convolutional layer for $T$ time steps results in a feed-forward subnetworks of depth $T+1$, as shown in Figure 3(c) where $T=2$. In this work, we set $T=2$. While the recurrent input evolves over iterations, the feed-forward input remains the same in all iterations. When $t=0$ only the feed-forward input is present. In the subnetwork, for a unit located at

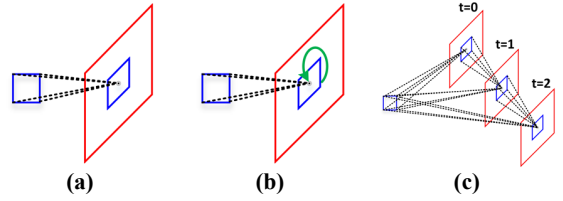

Figure 3. (a) Feed-forward convolutional layer. (b) Recurrent convolutional layer (RCL). (c) Subnetwork by unfolding the RCL.

$(i, j)$ on the $k$th feature map, its net input $z_{ijk}(t)$ at time step $t$ is computed as [43] by:

$$z_{ijk}(t) = \left(\mathbf{v}_k^f\right)^T \mathbf{u}^{(i,j)}(t) + (\mathbf{v}_k^r)^T \mathbf{x}^{(i,j)}(t-1) + b_k \quad (1)$$

where $\mathbf{u}^{(i,j)}(t)$ and $\mathbf{x}^{(i,j)}(t-1)$ denote the feed-forward and recurrent input, $\mathbf{v}_k^f$ and $\mathbf{v}_k^r$ denote the feed-forward weights and recurrent weights, and $b_k$ is the bias. More details about RCNN can be found in its reference [43]. As we can see, compared with the convolutional layer, the effective receptive field of an RCL unit in the feature maps of the previous layer expands when the iteration number increases. Therefore, the RCL is able to learn features which contain more contextual information.

To make the model learn more discriminative features from both local and global views, a side-output is generated to perform the deeply-supervised learning from the last RCL of each block by a convolutional layer and a deconvolutional layer, as shown in Figure 2. Another benefit of deeply-supervised learning is to alleviate the common problem of "vanishing" gradients during training such a deep network. The additional convolutional layer with one $1 \times 1$ convolutional kernel converts the feature maps to a salient map, and the additional deconvolutional layer is used to make the salient map have the same size with the input image. To make the final salient map consider the local and global information of salient objects and be robust to the size variation of salient objects, the side-outputs of the all five blocks are fused by concatenating them in the channel direction and using a convolutional kernel with size of $1 \times 1$ to convert the concatenation maps to the final salient map. Convolution with a $1 \times 1$ kernel is a weighted fusion process. So far, the whole architecture of DSRCNN has been constructed, as shown in Figure 2. All additional deconvolutional layer and the last convolutional layer are followed by a sigmoid activation function. And for each block, a dropout layer with the dropout ratio of 0.5 is followed to alleviate the overfitting.

For training of DSRCNN, the errors between all side-outputs and the ground truth should be computed and backward propagate. Therefore, we need to define a loss function to compute these errors. For most of images, the number of pixels in salient objects and background are heavily imbalanced. Here, given an image $X$ and its ground truth $Y$, a cross-entropy loss function defined in [40] is used to balance the loss between salient and background classes as follows:

$$l_{side}^m(\mathbf{W}, \mathbf{w}^m) = -\alpha \sum_{i=1}^{|Y_+|} \log P(y_i = 1|X; \mathbf{W}, \mathbf{w}^m) \\ -(1-\alpha) \sum_{i=1}^{|Y_-|} \log P(y_i = 0|X; \mathbf{W}, \mathbf{w}^m) \quad (2)$$

where $\alpha = |Y_-|/(|Y_+| + |Y_-|)$, $|Y_+|$ and $|Y_-|$ mean the number of salient pixels and background pixels in ground truth, $\mathbf{W}$ denotes the parameters of all network layers in the five blocks, $\mathbf{w}^m$ denotes the weights of the $m^{\text{th}}$ side-output layer including a

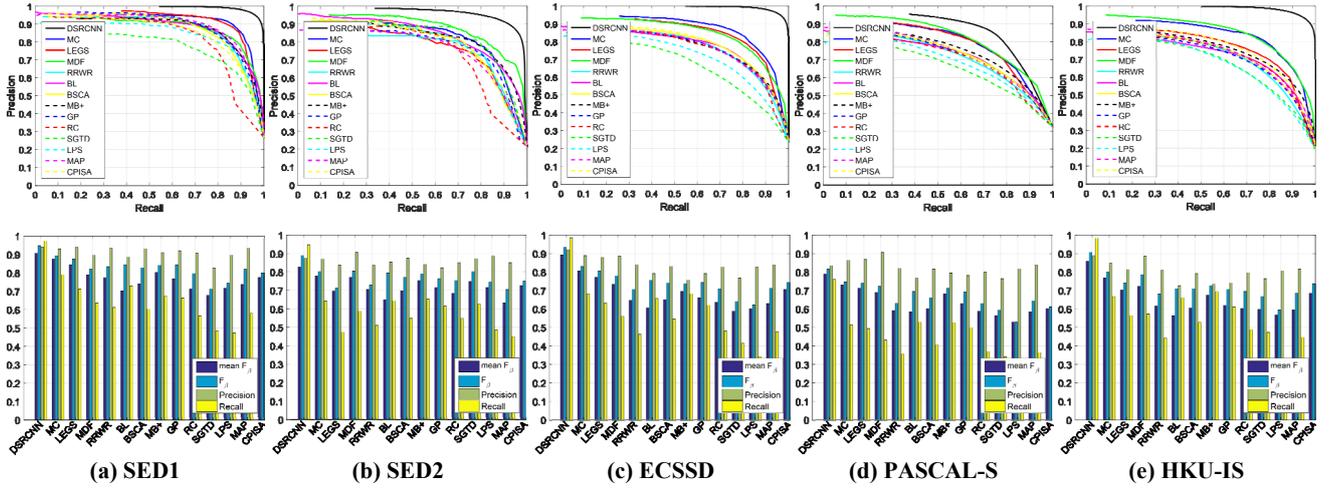

**(a) SED1**  **(b) SED2**  **(c) ECSSD**  **(d) PASCAL-S**  **(e) HKU-IS**

Figure 4: Results of all test approaches on five standard benchmark datasets, i.e. SED1, SED2, ECSSD, PASCAL-S, and HKU-IS. The first row presents the precision-recall curves. The second row presents the mean F-measures and the adaptive F-measures/precision/recall which are computed from the binary images obtained by binarizing the salient maps with adaptive thresholds computed by using Otsu algorithm.

Table 1. The weighted F-measure and MAE of different approaches on different test datasets (red, blue, and green texts respectively indicate rank 1, 2, and 3)

| Approach | Year | SED1 | | SED2 | | ECSSD | | PASCAL-S | | HKU-IS | |
|---|---|---|---|---|---|---|---|---|---|---|---|
| | | wF$_\beta$ | MAE | wF$_\beta$ | MAE | wF$_\beta$ | MAE | wF$_\beta$ | MAE | wF$_\beta$ | MAE |
| *DSRCNN* | / | 0.8935 | 0.0357 | 0.7924 | 0.0526 | 0.8718 | 0.0368 | 0.6974 | 0.1284 | 0.8330 | 0.0399 |
| MC | CVPR2015 | 0.8242 | 0.0790 | 0.6959 | 0.1153 | 0.7293 | 0.1019 | 0.6064 | 0.1422 | 0.6899 | 0.0914 |
| LEGS | CVPR2015 | 0.7671 | 0.1021 | 0.5911 | 0.1407 | 0.6722 | 0.1256 | 0.5791 | 0.1593 | 0.5911 | 0.1301 |
| MDF | CVPR2015 | 0.6896 | 0.1284 | 0.6674 | 0.1152 | 0.6194 | 0.1377 | 0.5386 | 0.1633 | 0.6135 | 0.1152 |
| RRWR | CVPR2015 | 0.6524 | 0.1412 | 0.5914 | 0.1615 | 0.5026 | 0.1850 | 0.4435 | 0.2262 | 0.4592 | 0.1719 |
| BL | CVPR2015 | 0.5612 | 0.1850 | 0.4673 | 0.1905 | 0.4615 | 0.2178 | 0.4464 | 0.2478 | 0.4119 | 0.2136 |
| BSCA | CVPR2015 | 0.6161 | 0.1546 | 0.5426 | 0.1591 | 0.5159 | 0.1832 | 0.4703 | 0.2220 | 0.4643 | 0.1760 |
| MB+ | ICCV2015 | 0.6956 | 0.1334 | 0.6354 | 0.1379 | 0.5632 | 0.1717 | 0.5307 | 0.1964 | 0.5438 | 0.1497 |
| GP | ICCV2015 | 0.6449 | 0.1536 | 0.5688 | 0.1620 | 0.5180 | 0.1919 | 0.4823 | 0.2300 | 0.4680 | 0.1852 |
| RC | TPAMI2015 | 0.6033 | 0.1642 | 0.5461 | 0.1562 | 0.5118 | 0.1868 | 0.4694 | 0.2253 | 0.4768 | 0.1714 |
| SGTD | TIP2015 | 0.5743 | 0.1855 | 0.6453 | 0.1285 | 0.4689 | 0.2007 | 0.4385 | 0.2269 | 0.4785 | 0.1627 |
| LPS | TIP2015 | 0.6029 | 0.1610 | 0.5950 | 0.1411 | 0.4585 | 0.1877 | 0.3882 | 0.2162 | 0.4252 | 0.1635 |
| MAP | TIP2015 | 0.6339 | 0.1459 | 0.5181 | 0.1703 | 0.4953 | 0.1861 | 0.4361 | 0.2222 | 0.4533 | 0.1717 |
| CPISA | TIP2015 | 0.6645 | 0.1458 | 0.5939 | 0.1482 | 0.5735 | 0.1596 | 0.4478 | 0.1983 | 0.5575 | 0.1374 |

convolutional layer and a deconvolutional layer, and $P(y_i = 1|X; \mathbf{W}, \mathbf{w}^m) \in [0,1]$ is computed using sigmoid function on the activation value at pixel $j$. So the whole loss of all side-outputs is defined as

$$L_{side}(\mathbf{W}, \mathbf{w}) = \sum_{m=1}^{5} l_{side}^m(\mathbf{W}, \mathbf{w}^m) \quad (3)$$

where $\mathbf{w} = (\mathbf{w}^1, \mathbf{w}^2, ..., \mathbf{w}^5)$. The loss between the fusion output and ground truth $L_{fuse}(\mathbf{W}, \mathbf{w}, \mathbf{h})$ is also computed by using the cross-entropy loss function, where $\mathbf{h}$ is the weights of the last fusion convolutional kernel. After computing all losses, the standard stochastic gradient descent algorithm is used to minimize the following objective function during training.

$$(\mathbf{W}, \mathbf{w}, \mathbf{h})^* = \text{argmin}\left(L_{side}(\mathbf{W}, \mathbf{w}) + L_{fuse}(\mathbf{W}, \mathbf{w}, \mathbf{h})\right) \quad (4)$$

For testing of DSRCNN, given an image, we can use the trained model to predict a salient map, in which the salient object regions have larger values.

We use the popular Caffe deep leaning library [45] to implement the framework of DSRCNN. The THUS-10000 dataset [34] which contains 10,000 images and their corresponding ground truths is used for training. Since DSRCNN is a fully convolutional network, the images with arbitrary sizes can be the inputs without being resized during training and test. The training and test processes are conducted on a PC with an Intel i7-4790k CPU, a TESLA K40c GPU, and 32G RAM. The average runtime per image on the THUS-10000 dataset is about 0.23 second.

## 3. EXPERIMENTAL RESULTS
### 3.1 Datasets and Evaluation Criteria

We evaluate the proposed method, denoted as DSRCNN, on five standard benchmark datasets: SED1 [46], SED2 [46], ECSSD [8], PASCAL-S [20], and HKU-IS [21].

**SED1** [46] and **SED2** [46] contain 100 images with one and two salient object, respectively, in which objects have largely different sizes and locations.

**ECSSD** [8] contains 1,000 images with complex backgrounds, which makes the detection tasks much more challenging.

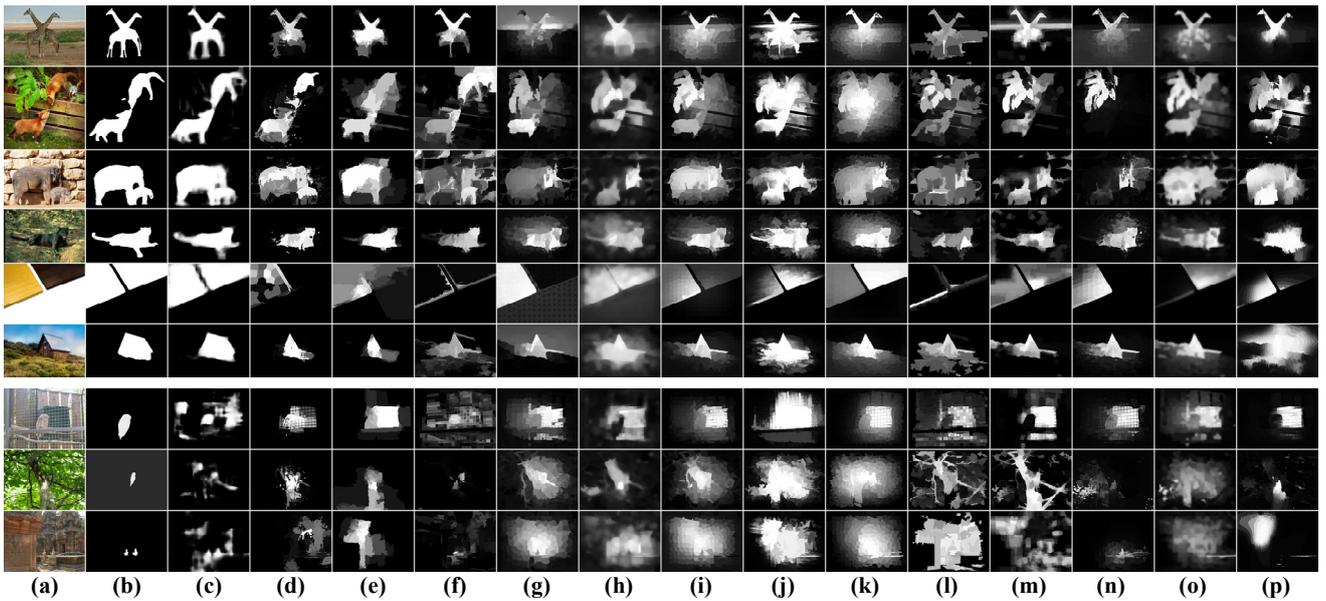

Figure 5: Visual Comparisons of different saliency detection approaches in various challenging scenarios. Our method is successful and failed in the first six rows and the last three rows, respectively. (a) Original images, (b) Ground truth, (c) DSRCNN, (d) MC, (e) LEGS, (f)MDF, (g) RRWR, (h) BL, (i) BSCA, (j) MB+, (k) GP, (l) RC, (m) SGTD, (n) LPS, (o) MAP, (p) CPISA.

**PASCAL-S** [20] is constructed on the validation set of the PASCAL VOC 2012 segmentation challenge. This dataset contains 850 natural images with multiple complex objects and cluttered backgrounds. The PASCAL-S data set is arguably one of the most challenging saliency data sets without various design biases (e.g., center bias and color contrast bias).

**HKU-IS** [21] contains 4447 challenging images, which is newly developed by considering at least one of the following criteria: 1) there are multiple disconnected salient objects, 2) at least one of the salient objects touches the image boundary, 3) the color contrast (the minimum Chi-square distance between the color histograms of any salient object and its surrounding regions) is less than 0.7.

All datasets provide the corresponding ground truth in the form of accurate pixel-wise human-marked labels for salient regions.

For evaluation, the popular criteria, i.e. the standard precision-recall (PR) curves, F-measure (denoted as $F_\beta$), the mean absolute error (MAE), and the weighted F-measure (denoted as $wF_\beta$) [47], are used to test the performance of the proposed method.

### 3.2 Performance Comparison

We compare the proposed method (denoted as DSRCNN) with thirteen state-of-the-art saliency detection approaches on five datasets, including MC [26], LEGS [28], MDF [21], RRWR [27], BL [23], BSCA [25], MB+ [30], GP [36], RC [34], SGTD [35], LPS [32], MAP [33], CPISA [31]. For fair comparison, the source codes of these approaches released by the authors are used for test with recommended parameter settings in this work.

Figure 4 shows the PR curve and $F_\beta$ of different approaches on all test datasets, and Table 1 lists the MAE and $wF_\beta$ of different approaches on all test datasets. From Figure 4 and Table 1, we can see that the proposed method significantly outperforms all of the state-of-the-art approaches on all test datasets in terms of all evaluation criteria, which convincingly demonstrates the effectiveness of the proposed method. All methods get the worst performance on the PASCAL-S dataset. That is because PASCAL-S is the most complex one in these five test datasets, and different salient objects get different saliencies in the ground truths. Therefore, we believe that if a large-scale dataset containing complex images and different saliencies for different salient objects is used to train the DSRCNN model, the performance will be further improved. Also, we qualitatively compare our detected salient maps with those detected by other approaches in the first six rows of Figure 5. As we can see, the proposed method is able to highlight saliencies of salient objects and suppress the saliencies of background better, and the results of our salient maps are much close to the ground truth in various challenging scenarios.

The last three rows of Figure 5 show some cases in which the proposed method fails to correctly detect the salient regions. For example, the colors of salient objects and backgrounds are very similar, the backgrounds are too complex, and the salient objects are too small. In these cases, the other approaches also cannot correctly detect the salient objects.

### 4. CONCLUSIONS

In this paper, we have developed a deeply-supervised recurrent convolutional neural network (DSRCNN) for saliency detection, which integrates the ideas of recurrent convolutional neural networks and deeply-supervised networks. DSRCNN not only learns the explicit contextual and high-level information, but also combines multi-scale discriminative information. All of these information is important for saliency detection, so the proposed DSRCNN based saliency detection method is able to generate high quality salient maps for various challenging scenarios. Extensive experiments on five standard benchmark datasets demonstrate that the proposed method get the best performance compared with the state-of-the-art approaches, convincingly demonstrating the effectiveness of the proposed method.

### 5. ACKNOWLEDGEMENTS

This work was supported by the Natural Science Foundation of China under Grant 61472102. The authors would like to thank the founders of the publicly available datasets and the support of NVIDIA Corporation with the donation of the Tesla K40 GPU used for this research.